\documentclass[journal]{IEEEtran}

\ifCLASSINFOpdf

\else

\fi

\usepackage{graphicx,wrapfig,fullpage,amsmath,hhline,epsfig,verbatim,url,amssymb,multicol,multirow,cite}
\usepackage{nomencl}

\usepackage{times,color,soul}
\usepackage[left=0.75in,top=0.70in,right=0.75in,bottom=0.80in]{geometry}
\usepackage{amsbsy,latexsym,mathrsfs,mathtools}
\makenomenclature
\usepackage{mathptmx}
\DeclareMathAlphabet{\mathcal}{OMS}{cmsy}{m}{n}
\usepackage{bm}
\usepackage{float}

\usepackage{color,soul}
\usepackage{dsfont}
\usepackage{comment}
\usepackage{algorithm}
\usepackage{algorithmic}
\usepackage{psfrag}   

\usepackage{url}
\usepackage{hyperref}
\usepackage{setspace}
\usepackage[table]{xcolor}
\usepackage{paralist}
\usepackage{booktabs}
\usepackage{gensymb}
\usepackage{flushend}

\usepackage{hyperref}

\usepackage[switch]{lineno}

\begin{document}

\title{Invariant Filtering for Legged Humanoid Locomotion on a Dynamic Rigid Surface}

\author{Yuan~Gao,~\IEEEmembership{Student Member,~IEEE,}
        Chengzhi~Yuan,~\IEEEmembership{Member,~IEEE,}
        Yan~Gu,~\IEEEmembership{Member,~IEEE}
\thanks{
This work was supported by the National Science Foundation under Grants CMMI-1934280 and CMMI-2046562.
Y. Gao is with the Department of Mechanical Engineering, University of Massachusetts Lowell, Lowell, MA 01854, U.S.A.
Y. Gu is with the School of Mechanical Engineering, Purdue University, West Lafayette, IN 47907, U.S.A.
This work was partly conducted while Y. Gu was with the University of Massachusetts Lowell.
C. Yuan is with the Department of Mechanical, Industrial and Systems Engineering, University of Rhode Island, Kingston, RI 02881, U.S.A.
Emails: {yuan\_gao}@student.uml.edu, cyuan@uri.edu, {yan.gu.purdue}@gmail.com.}}%

\maketitle

\begin{abstract}

State estimation for legged locomotion over a dynamic rigid surface (DRS), which is a rigid surface moving in the world frame (e.g., ships, aircraft, and trains), remains an under-explored problem.
This paper introduces an invariant extended Kalman filter that
estimates the robot's pose and velocity during DRS locomotion by using common sensors of legged robots (e.g., inertial measurement units (IMU), joint encoders, and RDB-D camera).
A key feature of the filter lies in that it explicitly addresses the nonstationary surface-foot contact point and the hybrid robot behaviors.
Another key feature is that, in the absence of IMU biases, the filter satisfies the attractive group affine and invariant observation conditions,
and is thus provably convergent for the deterministic continuous phases.
The observability analysis
is performed to reveal the effects of DRS movement on the state observability, and the convergence property of the hybrid, deterministic filter system is examined for the observable state variables.
Experiments of a Digit humanoid robot walking on a pitching treadmill validate the effectiveness of the proposed filter under large estimation errors and moderate DRS movement.
The video of the experiments can be found at: \href{https://youtu.be/ScQIBFUSKzo}{https://youtu.be/ScQIBFUSKzo}.
\end{abstract}


\IEEEpeerreviewmaketitle

\section{Introduction}

\label{section: introduction}

State estimation is essential to providing the estimates of a robot's movement state (e.g., pose and velocity) needed for planning and control.
While state estimation for locomotion on static~\cite{bloesch2013state}
or relatively unstable surfaces~\cite{kim2021legged}
has been extensively studied, 
state estimation for dynamic rigid surfaces (DRS), such as ships and aircraft~\cite{iqbal2020provably}, has not been fully investigated.
{This paper aims to solve the estimation problem for DRS locomotion.} 
Yet, solving this problem is challenging due to the nonstationary surface-foot contact point
~\cite{iqbal2020provably,iqbal_SLIP} 
and the hybrid robot dynamics involving both continuous behaviors and discrete foot-landing events~\cite{gao2019dscc,gu2018exponential,8815144}.

Extended Kalman filtering (EKF)~\cite{lin2006sensor, fallon2014drift}
has been used to achieve real-time state estimation of legged locomotion on static surfaces by fusing the data returned by common on-board sensors such as encoders (which measure the joint angles) as well as IMUs attached to the robot's base (which measure the base's linear acceleration and angular velocity in the base frame).
Recently, EKF-based estimators have been created to estimate a robot's base pose and velocity
~\cite{bloesch2013state,ramadoss2021diligent}.
These methods can be applied to general legged locomotion because their formulation is independent of robot dynamics and gait types.
Specifically, the process model includes the IMU motion and bias dynamics, and the measurement model is based on the the leg odometry formed via the forward kinematics between the base and the ground contact point.
Yet, they may not be effective for DRS locomotion because they assume the foot-ground contact point is static in the world frame.
Also, 
similar to standard EKF, they may not handle large estimation errors well {because the underlying system linearization depends on the true state but is evaluated at the state estimate}~\cite{7523335}.

To ensure provable, rapid convergence under large estimation error, the previous EKF-based design~\cite{bloesch2013state} has been transformed into an invariant extended Kalman filter (InEKF) for legged locomotion on static surfaces~\cite{hartley2020contact,lin2021deep}.
{By the theory of InEKF~\cite{barrau2015non}, when the InEKF system meets the group affine and invariant observation conditions, 
the system linearization is independent of the true state and thus is valid even under relatively large errors.}
Still, the effectiveness of these methods in handling DRS locomotion is unclear, especially under a relatively significant surface motion (e.g., ship motion under sea waves~\cite{kuchler2011heave}), because of the underlying assumption of stationary surface-foot contact.

To address hybrid robot behaviors, state estimators for hybrid models of legged locomotion ~\cite{hamed2018observer,KONG2021109752} have been derived.
However, the convergence property of {invariant filters} for hybrid locomotion models has not been examined.

This paper introduces an InEKF method to produce real-time, accurate state estimation for bipedal humanoid walking on a DRS even under relatively large estimation errors. 
The main contributions are:
\begin{itemize}     

    \item [(a)] Deriving an InEKF that considers the DRS motion and hybrid robot behaviors and {meets} the group affine and invariant observation properties
    without IMU biases.
    \item [(b)] Building a {right-invariant} measurement model based on the rotational constraint at the surface-foot contact area, enhancing the convergence rate and rendering the {base yaw angle} observable under general DRS movement.
    \item [(c)] Performing observability analysis
    that reveals how the DRS pose affects state observability, and proving the stability for the hybrid, deterministic error dynamics.
    \item [(d)] Demonstrating the computational efficiency, accuracy, and robustness of the proposed filter through experiments of humanoid walking on a rocking treadmill.
\end{itemize}
{Some results in this paper have been reported~\cite{MECC_yuan}.
The new, substantial contributions of this paper are the last three items in the aforementioned list.}

The paper is structured as follows.
Section~\ref{Section-math background} provides a brief background on matrix Lie groups.
Section~\ref{section: problem formulation} presents the problem formulation.
Section~\ref{section: InEKF} introduces the proposed InEKF for the hybrid model of DRS locomotion.
Section~\ref{section: observability analysis} provides observability and convergence analysis.
Section~\ref{section: experiments} reports experiment results.
Section~\ref{section: discussion} discusses the capabilities and limitations of the filter.
Section~\ref{section: conclusion} gives the concluding remarks.

\section{PRELIMINARIES}

\label{Section-math background}
The matrix Lie group, denoted as $\mathcal{G}$, is a subset of $ n \times n$ invertible square matrices.
The associated Lie algebra $\boldsymbol{\mathfrak{g}}$ with a dimension of $\mbox{dim}\boldsymbol{\mathfrak{g}}$ is a set of $ n \times n$ square matrices.
The linear operator $(\cdot)^\wedge $ maps any vector $\boldsymbol{\xi}\in\mathbb{R}^{\text{dim} \boldsymbol{\mathfrak{g}}}$ onto $\boldsymbol{\mathfrak{g}}$.
The exponential map, $\text{exp}:\mathbb{R}^{\text{dim}\boldsymbol{\mathfrak{g}}}\rightarrow\mathcal{G}$, is defined as:
$\text{exp}(\boldsymbol{\xi}) \triangleq \text{expm}(\boldsymbol{\xi}^\wedge)$,
where \text{expm} is the usual exponential of $n \times n$ matrices. 
The inverse operator of $ (\cdot)^\wedge$ is denoted as $(\cdot)^\vee:\boldsymbol{\mathfrak{g}}\rightarrow \mathbb{R}^{\text{dim}\boldsymbol{\mathfrak{g}}}$.
The adjoint matrix $\mathbf{Ad}_{X}$ at $\mathbf{X}$ for any vector $\boldsymbol{\xi} \in\mathbb{R}^{\text{dim}\boldsymbol{\mathfrak{g}}}$ is defined as
$\mathbf{Ad}_{X}\boldsymbol{\xi}=(\mathbf{X}\boldsymbol{\xi}^\wedge\mathbf{X}^{-1})^\vee$.
More detailed introductions to matrix Lie groups can be found in~\cite{sola2018micro}.
A nomenclature table is given in supplementary material.

\section{PROBLEM FORMULATION}

\label{section: problem formulation}

\begin{figure}[t]
    \centering
    \includegraphics[width=0.65\linewidth]{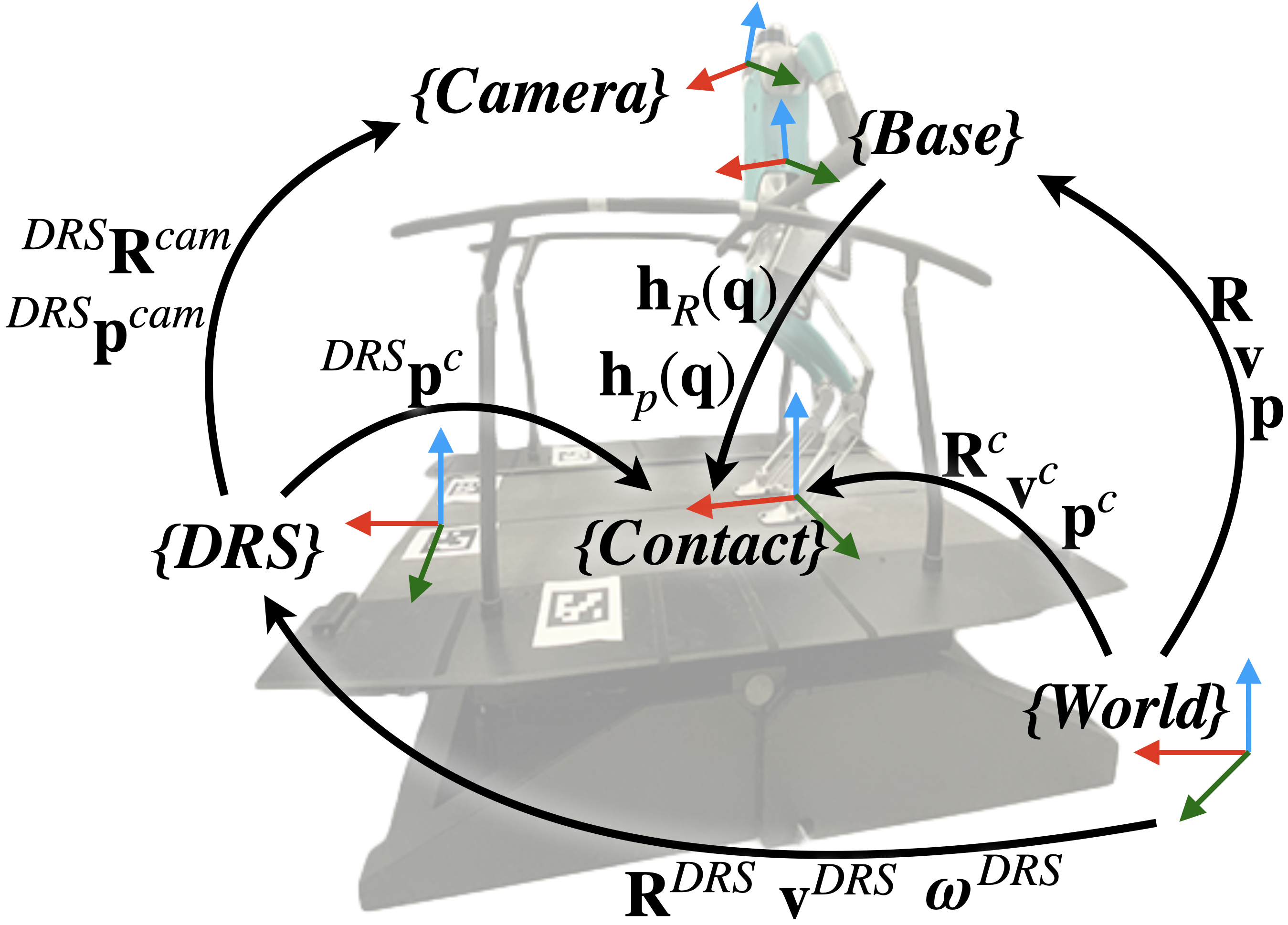}
    \caption{Illustration of coordinate frames and key variables. {The treadmill is a DRS that rotates in the world frame.}
    }
    \label{Fig: overview}
\end{figure}

{In the proposed filter design,} the state is chosen as {variables that are often needed in locomotion planning and control}, including the linear velocity $\mathbf{v} \in \mathbb{R}^3$ and orientation $\mathbf{R} \in SO(3)$ of the robot's base (e.g., chest) expressed in the world frame (see Fig.~\ref{Fig: overview}).
The state also includes the base position $\mathbf{p} \in \mathbb{R}^3$ and contact-point position $\mathbf{p}^{c} \in \mathbb{R}^3$ (see {Fig.~\ref{Fig: overview}}), so as to exploit the forward kinematics between the base and contact/foot frames in the filter design~\cite{bloesch2013state,hartley2020contact}.

The DRS of interest possesses two common characteristics of real-world DRSes such as aircraft and vessels.
First, when traveling on such surfaces, a robot can only see landmarks attached to the surface instead of the world frame, e.g., due to the concealed environment on the surface.
Second, the DRS orientation $\mathbf{R}^{DRS} \in SO(3)$ and linear and angular velocities $\mathbf{v}^{DRS}, \boldsymbol{\omega}^{DRS} \in \mathbb{R}^3$ (see Fig.\ref{Fig: overview}) are relatively accurately known.
This is a reasonable assumption because these real-world DRSes are typically equipped with high-accuracy motion monitoring systems~{\cite{vectorNav}}.
Also, the proposed filter design explicitly treats the inaccurate knowledge of surface pose and motion as explained in Sec.~\ref{section: InEKF}.

The sensors considered in this study are common on-board sensors, which are: an IMU attached to the robot's base, joint encoders, a RGB-D camera, and a contact indicator. 
The RGB-D camera tracks the landmarks attached to the DRS, which is used to obtain the camera pose in the DRS frame.
The contact indicator detects foot landing events.

The encoders measure the joint angles
$\mathbf{q} \in \mathbb{R}^{m}$
with $m$
the number of joints.
Corrupted by white {zero-mean} Gaussian noise $\mathbf{w}^q$,
the raw encoder data $\mathbf{\tilde{q}}$ is expressed as: 
$\mathbf{\tilde{q}} = \mathbf{q}  +  \mathbf{w}^q$.

The IMU includes a gyroscope and an accelerator that respectively measure the angular velocity $\boldsymbol{\omega}\in \mathbb{R}^3$ and linear acceleration $\mathbf{a}\in \mathbb{R}^3$ of the IMU in the base frame.
Corrupted by white Gaussian zero-mean noise $\mathbf{w}^a,\mathbf{w}^\omega \in \mathbb{R}^3$, as well as biases $\mathbf{b}^a,\mathbf{b}^\omega \in \mathbb{R}^3$, the IMU readings $\tilde{\mathbf{a}}$ and $\tilde{\boldsymbol{\omega}}$ are expressed as:
$\tilde{\mathbf{a}} = \mathbf{a}+\mathbf{b}^{a}+\mathbf{w}^a$
and
$\tilde{\boldsymbol{\omega}} =\boldsymbol{\omega} + \mathbf{b}^\omega + \mathbf{w}^\omega$.

\subsection{Continuous-Phase IMU Motion and Bias Dynamics}
To form the process model, we choose to adopt the IMU motion dynamics due to its accuracy and simplicity~\cite{bloesch2013state}.
At time $t$, the IMU motion dynamics is given by:
\begin{equation}
    \begin{gathered}
        \dot{\mathbf{R}}_t = \mathbf{R}_t(\boldsymbol{\tilde{\omega}}_t-\mathbf{b}_t^\omega-\mathbf{w}_t^\omega)_{\times},
        \\
        \dot{\mathbf{v}}_t = \mathbf{R}_t(\tilde{\mathbf{a}}_t-\mathbf{b}^a_t-\mathbf{w}_t^a)+\mathbf{g},
        ~\mbox{and}~
         \dot{\mathbf{p}}_t = \mathbf{v}_t,
    \end{gathered}
    \label{equ: IMU motion dynamics}
\end{equation}
where
$ (\cdot)_\times$ is a skew-symmetric matrix
and
$\mathbf{g}$ is the gravitational acceleration {vector}.
The IMU bias dynamics is modeled as Brownian motion~\cite{hartley2020contact}:
\begin{equation}
    \begin{aligned}
        \dot{\mathbf{b}}^a_t = \mathbf{w}_t^{ba} 
        ~\mbox{and}~
        \dot{\mathbf{b}}^\omega_t = 
        \mathbf{w}_t^{b\omega}, 
    \end{aligned}
    \label{equ: bias dynamics}
\end{equation}
where $\mathbf{w}_t^{ba}$ and $\mathbf{w}_t^{b\omega}$ are white zero-mean Gaussian noise.

\subsection{Continuous-Phase Contact-Point Motion Dynamics}

During DRS locomotion, the foot moves in the world frame due to the surface movement.
Thus, the deterministic motion model of the contact point is not $\dot{\mathbf{p}}^c_t=\mathbf{0}$ as in previous work~\cite{bloesch2013state,hartley2020contact} on static surfaces.
Instead, we explicitly consider the contact point velocity $\mathbf{v}^c_t$ in the model:
\begin{equation}
    \dot{\mathbf{p}}^c_t=\mathbf{v}^c_t.
    \label{equ: contact point dynamics}
\end{equation}

{In this study, to inform the model in Eq.~\eqref{equ: contact point dynamics}, we choose to directly measure the contact point velocity based on the known surface pose and motion and the measured contact position in the DRS frame through the} following kinematics:
\begin{equation}
    \mathbf{{v}}_t^c = \mathbf{v}_t^{DRS}+\boldsymbol{\omega}^{DRS}_t\times (\mathbf{R}_t^{DRS} {~^{DRS}\mathbf{{p}}_t^{c})}.
    \label{eqn: vd}
\end{equation}
Here,
$^{DRS}\mathbf{{p}}_t^{c}$ is the contact point position relative to the DRS frame, expressed in the DRS frame (see Fig.~\ref{Fig: overview}).
Note that the computation of the velocity $ ^{DRS}\mathbf{{p}}_t^{c}$ depends on the robot's camera data and the joint angle data returned by encoders.
Also, recall that the surface orientation $\mathbf{R}_t^{DRS}$ and motion  $\boldsymbol{\omega}^{DRS},\mathbf{v}_t^{DRS}$ are assumed to be known as explained earlier.
An example of computing $\mathbf{v}^c_t$ is given in Sec.~\ref{section: experiments}.

The velocity computation inaccuracy is considered as:
\begin{equation}
    \mathbf{\tilde{v}}_t^c = \mathbf{v}_t^c +\mathbf{R}_t {\mathbf{w}}_t^c,
    \label{equ: contact point noise}
\end{equation}
where $\tilde{\mathbf{v}}^c_t \in \mathbb{R}^{3}$ is the measured contact point velocity, and the inaccuracy $\mathbf{w}_t^c$ is modeled as white Gaussian zero-mean noise expressed in the base frame.

\subsection{Discrete Jump Dynamics at a Foot Landing}

At a foot landing, the swing and support legs switch roles, 
causing a discrete jump in the contact point position $\mathbf{p}^c_t$.
To appropriately propagate the estimate and covariance at foot landings, we choose to explicitly consider the jump.

The jump map of the contact point position $\mathbf{p}^c_t$ is:
\begin{equation}
    \mathbf{p}^c_{t^+} = \mathbf{p}^c_{t} + \mathbf{R}_{t} \mathbf{h}_{c}(\mathbf{q}_{t})
    \label{equ: pc jump}
\end{equation}
where the subscript $t^+$ denotes the timing just after the foot landing at $t$.
Here the function $\mathbf{h}_c$ is the forward kinematics from the previous
support-foot position to the new one, expressed in the base frame.
Except for $\mathbf{p}^c_{t}$, all other state variables remain continuous at foot switching.

With the first-order Taylor expansion,
the nonlinear term in the jump dynamics (Eq.~\eqref{equ: pc jump}) can be approximated as:
$\mathbf{R}_{t} \mathbf{h}_{c}(\mathbf{q}_{t}) 
\approx
\mathbf{R}_{t} \mathbf{h}_{c}(\tilde{\mathbf{q}}_{t}) 
-
\mathbf{R}_{t}
\frac{\partial \mathbf{h}_c}{\partial \mathbf{q}} (\mathbf{\Tilde{q}}_{t}) \mathbf{w}^{q}_t$.

\subsection{Position based Forward Kinematics Measurement}

\begin{figure}[t]
    \centering
    \includegraphics[width=0.8\linewidth]{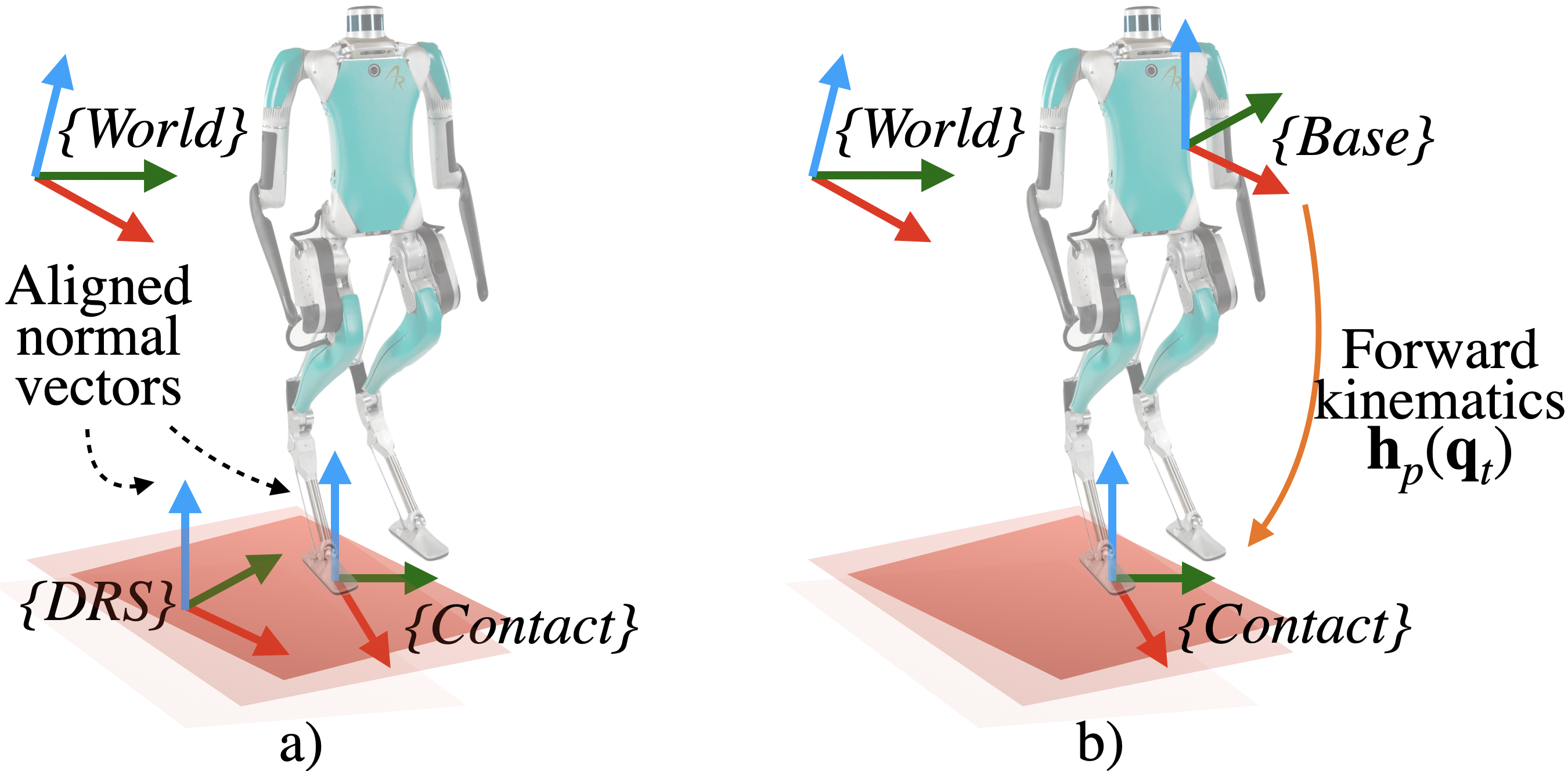}
    \caption{{Illustrations} of the observations: a) normal vector alignment of the contact and DRS frames and b) contact point position in the base frame.
    }
    \label{Fig:invariant_measurement_model}
\end{figure}

To connect the contact and the base frames, 
we adopt the leg odometry measurement in~\cite{bloesch2013state,hartley2020contact} (see Fig.~\ref{Fig:invariant_measurement_model}-b):
\begin{equation}
\label{equ: position measuremment}
    \mathbf{R}_{t}^T(\mathbf{p}^c_{t}-\mathbf{p}_{t})
    =
     \mathbf{h}_p(\mathbf{{q}}_{t}),
\end{equation}
where the forward kinematics function $\mathbf{h}_p$ is the support foot position relative to the base expressed in the base frame.
{Given the inaccuracy of the encoder reading $\tilde{\mathbf{q}}_{t}=\mathbf{q}  +  \mathbf{w}^q$ and with the first-order Taylor expansion, the model in Eq.~\eqref{equ: position measuremment} can be rewritten as:} 
$\mathbf{h}_{p}(\mathbf{q}_{t}) 
\approx  
\mathbf{h}_{p}(\tilde{\mathbf{q}}_{t}) - 
\frac{\partial \mathbf{h}_p}{\partial {\mathbf{q}}} (\mathbf{\Tilde{q}}_{t}) \mathbf{w}^{q}_t$.

\subsection{Contact Orientation based Measurement}

When the support foot and the surface have a full area contact, their normal vectors are parallel, whether the surface is stationary or moving (see Fig.~\ref{Fig:invariant_measurement_model}-a)).
{In this study, we utilize this rotational kinematic relationship to form a measurement model.}
Suppose that the $z$-axes of the contact and surface frames are aligned and normal to the DRS.
Then,
\begin{equation}
\label{equ: alligned normal vector}
\mathbf{R}^{DRS}_{t}
\begin{bmatrix}
    0~
    ~
    0
    ~
    1
\end{bmatrix}^T
=
{\mathbf{R}}^c_{t}
\begin{bmatrix}
    0
    ~
    0
    ~
    1
\end{bmatrix}^T
=
{\mathbf{R}}_{t}
\mathbf{h}_R(\mathbf{q}_t)
\begin{bmatrix}
    0
    ~
    0
    ~
    1
\end{bmatrix}^T
\end{equation}
holds, where $\mathbf{R}^c_{t} \in SO(3)$ is the contact frame orientation and the forward kinematics matrix function $\mathbf{h}_R$ is the support foot orientation with respect to the base frame (see Fig.\ref{Fig: overview}).

To address the inaccuracy of the known surface orientation $\mathbf{\tilde{R}}^{DRS}_{t}$, we assume the true orientation is corrupted by white Gaussian zero-mean uncertainty $\mathbf{w}^{DRS}_{t}$ as:
${\mathbf{R}}^{DRS}_{t}
    =
    \text{exp}(- \mathbf{w}^{DRS}_{t}) {\mathbf{\tilde{R}}}^{DRS}_{t}
    \approx
    (\mathbf{I}_3 -{(\mathbf{w}^{DRS}_{t})_\times)}  \tilde{\mathbf{R}}^{DRS}_{t}$,
where $\mathbf{I}_n$ is an $n \times n$ identity matrix.

To handle the inaccuracy of the encoder reading $\tilde{\mathbf{q}}_{t}$,
the support foot orientation
$\mathbf{R}_{t}  \mathbf{h}_R(\mathbf{q}_{t})$
is
approximated as:
$\mathbf{R}_{t}  \mathbf{h}_R(\mathbf{q}_{t})
    \approx
    \mathbf{R}_{t}\mathbf{h}_R(\mathbf{\tilde{q}}_{t})-\mathbf{R}_{t}\mathbf{J}_{h_R}(\tilde{\mathbf{q}}_{t},\mathbf{w}^q_{t})$.
Here the matrix $\mathbf{J}_{h_R}(\tilde{\mathbf{q}}_{t},\mathbf{w}^q_{t})$ is obtained based on the Jacobian of each column of $\mathbf{h}_R\triangleq[\mathbf{h}_{R,1},~\mathbf{h}_{R,2},~\mathbf{h}_{R,3}]$ as:
$
     \mathbf{J}_{h_R}
    \triangleq
    \begin{bmatrix}
        \frac{\partial \mathbf{h}_{R,1}}{\partial {\mathbf{q}}_{t}} (\tilde{\mathbf{q}}_{t_n}) \mathbf{w}^q_{t},
        ~
        \frac{\partial \mathbf{h}_{R,2}}{\partial {\mathbf{q}}_{t}}
        (\tilde{\mathbf{q}}_{t})
        \mathbf{w}^q_{t},
        ~
        \frac{\partial \mathbf{h}_{R,3}}{\partial {\mathbf{q}}_{t}}
        (\tilde{\mathbf{q}}_{t})
        \mathbf{w}^q_{t}
    \end{bmatrix}.
$

Combining these equations yields:
\begin{equation}
\label{equ:haha}
\begin{aligned}
&\mathbf{R}_{t}^T \mathbf{\tilde{R}}^{DRS}_{t}
\begin{bmatrix}
    0
    ~
    0
    ~
    1
\end{bmatrix}^T
+
\mathbf{R}^T_{t}(-\mathbf{w}^{DRS}_{t})_\times \mathbf{\tilde{R}}^{DRS}_{t}
\begin{bmatrix}
    0
    ~
    0
    ~
    1
\end{bmatrix}^T
\\
\approx &
\mathbf{h}_R(\mathbf{\tilde{q}}_{t})
{\begin{bmatrix}
    0
    ~
    0
    ~
    1
\end{bmatrix}^T}
-
\tfrac{\partial \mathbf{h}_{R,3}}{\partial \tilde{\mathbf{q}}_{t}}(\tilde{\mathbf{q}}_{t})
\mathbf{w}^q_{t}.
\end{aligned}
\end{equation}

\section{FILTER DESIGN}
\label{section: InEKF}

This section introduces the proposed InEKF design {based on the models formulated in Sec.~\ref{section: problem formulation}}.

The proposed filter derivation begins with proper state representation.
We adopt the representation in~\cite{hartley2020contact} since our filters estimate the same state.
First, the state variables
{$ \mathbf{R}_t$},
$\mathbf{v}_t$,
$\mathbf{p}_t$, and
$\mathbf{p}^c_{t}$
are expressed on the matrix Lie group
$\mathcal{G}$ as:
\begin{equation}
\label{equ: estimation state}
    \mathbf{X}_t \triangleq
    \begin{bmatrix}
    \mathbf{R}_t & \left[\mathbf{v}_t, ~ \mathbf{p}_t, ~ \mathbf{p}^c_{t} \right]
    \\
    \mathbf{0}_{3 \times 3} &    \mathbf{I}_3 
    \end{bmatrix}
    \in \mathcal{G},
\end{equation}
where $\mathbf{0}_{m \times n}$ is an $m \times n$ zero matrix.
The Lie group $\mathcal{G}$ is $SE_3(3)$, an extension of the special Euclidean group $SE(3)$.

To explicitly handle IMU biases, they are also chosen as state variables.
These biases are typically expressed on the vector space instead of $\mathcal{G}$~\cite{barrau2015non}; that is,
$\boldsymbol{\theta}_t \triangleq [(\mathbf{b}^{\omega}_t)^T,(\mathbf{b}^a_t)^T]^T$.

Let $\mathbf{\bar{(\cdot)}}$ denote the estimate of the variable $\mathbf{{(\cdot)}}$.
Based on the InEKF framework~\cite{7523335},
we use the right-invariant error $\boldsymbol{\eta}_t$
to represent the estimation error of $\mathbf{X}_t$ on $\mathcal{G}$:
\begin{equation}
    \boldsymbol{\eta}_t \triangleq
    \bar{\mathbf{{X}}}_t\mathbf{X}_t^{-1}
    \in \mathcal{G}.
\end{equation}
The log of the invariant error, denoted as $\boldsymbol{\xi}_t$, is a vector on $\mathbb{R}^{\text{dim}\boldsymbol{\mathfrak{g}}}$ defined via $\boldsymbol{\eta}_t \triangleq \text{exp}(\boldsymbol{\xi}_t)=\text{expm}(\boldsymbol{\xi}_t^{\wedge})$.
The expressions of $\boldsymbol{\eta}_t$, $\boldsymbol{\xi}_t$, and $\boldsymbol{\xi}_t^{\wedge}$ are in Sec. 2 of supplementary document.
The IMU bias error $\boldsymbol{\zeta}_t$ is defined as: 
$
    \boldsymbol{\zeta}_t 
    \triangleq
    \boldsymbol{\bar{\theta}}_t-\boldsymbol{\theta}.
$

\subsection{Continuous-Phase Process Model and Propagation Step}

This subsection introduces the process model and propagation step of the proposed filter for the continuous phases.

\subsubsection{Process model}
Based on the IMU motion and bias dynamics and the contact point motion in Eqs.~\eqref{equ: IMU motion dynamics}-\eqref{equ: contact point dynamics},
the process model
is expressed as:
\begin{equation}
\label{equ: process model}
    \begin{aligned}
    \dot{\mathbf{X}}_t &=
    \begin{bmatrix}
    \mathbf{R}_t(\Tilde{\boldsymbol{\omega}}_t-\mathbf{b}_t^\omega)_\times 
    & 
    \left[
    \mathbf{R}_t(\Tilde{\mathbf{a}}_t-\mathbf{b}_t^a) + \mathbf{g}, ~ \mathbf{v}_t,~ \tilde{\mathbf{v}}^{c}_t
    \right]
    \\
    \mathbf{0}_{3 \times 3} & \mathbf{0}_{3 \times 3}
    \end{bmatrix}
    \\
    &~~~~-\mathbf{X}_t(\mathbf{w}^X_t)^{\wedge}
    \triangleq
    \mathbf{f}_{u_t}(\mathbf{X}_t,\boldsymbol{\theta}_t)-\mathbf{X}_t (\mathbf{w}^X_t)^{\wedge},
    \end{aligned}
\end{equation}
with the noise vector
$
\mathbf{w}^X_t \triangleq
\begin{bmatrix}
    (\mathbf{w}^\omega_t)^T, (\mathbf{w}^a_t)^T, \mathbf{0}_{1 \times 3}, (\mathbf{w}^c_t)^T
\end{bmatrix}^T.$
Here we define the input $\mathbf{u}_t$ to consist of the IMU and encoder readings and the measured contact point velocity, i.e., $\mathbf{u}_t = \begin{bmatrix}
    \boldsymbol{\tilde{\omega}}_t^T, ~ \mathbf{\tilde{a}}_t^T, ~
    (\mathbf{\tilde{v}}_t^c)^T,~\tilde{\mathbf{q}}_t
\end{bmatrix}^T$.
Note that the encoder reading $\tilde{\mathbf{q}}_t$ is not an input to the continuous-phase process model in Eq.~\eqref{equ: process model} but is used later in the jump process model.

\subsubsection{Linearized error model}

By using the first-order Taylor expansion
$\boldsymbol{\eta}_t \approx \mathbf{I}+\boldsymbol{\xi}_t^\wedge$ and by applying the chain rule to express $\dot{\boldsymbol{\eta}}_t$, we obtain
the linearized error equation:
\begin{equation}
    \begin{bmatrix}
    \dot{\boldsymbol{\xi}}_t
    \\
    \dot{\boldsymbol{\zeta}}_t
    \end{bmatrix}
    =
    \mathbf{A}_t
    \begin{bmatrix}
    \boldsymbol{\xi}_t
    \\
    \boldsymbol{\zeta}_t
    \end{bmatrix}
    +
    \begin{bmatrix}
    \mathbf{Ad}_{\mathbf{\bar{X}}_t} &\mathbf{0}_{12\times6}
    \\
    \mathbf{0}_{6\times 12} &\mathbf{I}_{6}
    \end{bmatrix}
    \mathbf{w}_t.
    \label{equ: log-linear eqn}
\end{equation}
Here,
the noise term $\mathbf{w}_t$ is
$\mathbf{w}_t \triangleq [(\mathbf{w}_t^X)^T,
(\mathbf{w}_t^{b\omega})^T, (\mathbf{w}_t^{ba})^T]^T$,
the adjoint matrix
$\mathbf{Ad}_{\bar{\mathbf{X}}_t}$ is given in Sec. 3 of supplementary material, and the matrix $\mathbf{A}_t$ is:
\begin{equation}
\small
    \mathbf{A}_t=
    \begin{bmatrix}
        \mathbf{0}_{3\times 3} & \mathbf{0}_{3\times 3} &\mathbf{0}_{3\times 3} &
        \mathbf{0}_{3\times 3} & -\mathbf{\bar{R}}_t &\mathbf{0}_{3\times 3}
        \\
      (\mathbf{g})_{\times} & \mathbf{0}_{3\times 3} &\mathbf{0}_{3\times 3} &
        \mathbf{0}_{3\times 3}
        & -(\mathbf{\bar{v}}_t)_\times\mathbf{\bar{R}}_t &-\mathbf{\bar{R}}_{t}
        \\
        \mathbf{0}_{3\times 3} & \mathbf{I}_{3} &\mathbf{0}_{3\times 3} &
        \mathbf{0}_{3\times 3}
        & -(\mathbf{\bar{p}}_t)_\times\mathbf{\bar{R}}_t &\mathbf{0}_{3\times 3}
        \\
        (\Tilde{\mathbf{v}}_{t}^c)_{\times} & \mathbf{0}_{3\times 3} &\mathbf{0}_{3\times 3} &
        \mathbf{0}_{3\times 3}
        & -(\mathbf{\bar{p}}^c_t)_\times\mathbf{\bar{R}}_t &\mathbf{0}_{3\times 3}
        \\
        \mathbf{0}_{6\times3} & \mathbf{0}_{6\times3}&\mathbf{0}_{6\times3} & \mathbf{0}_{6\times3} &
        \mathbf{0}_{6\times3} &
        \mathbf{0}_{6\times3}
    \end{bmatrix}.
    \label{equ: A}
\end{equation}
Note that $\mathbf{A}_t$
contains the contact point velocity $\mathbf{\tilde{v}}_t^c$
because the process model explicitly considers it.
Derivation of Eqs.~\eqref{equ: log-linear eqn} and \eqref{equ: A} is in Sec. 3 of supplementary material.

\subsubsection{Propagation}

Let $t_n$ ($n \in \{1,2,...\}$) denote the time when sensors return data for estimation error correction.
Then, during the propagation step on $t \in [t_{n-1}, t_{n})$,
the estimates $\bar{\mathbf{X}}_t$ and $\bar{\boldsymbol{\theta}}_t$ are obtained via $\dot{\bar{\mathbf{X}}}_t=\mathbf{f}_{u_t}(\bar{\mathbf{X}}_t,\bar{\boldsymbol{\theta}}_t)$
and $\dot{\bar{\boldsymbol{\theta}}}_t = \mathbf{0}$
based on the process models in Eqs.~\eqref{equ: process model} and \eqref{equ: bias dynamics}.

By the InEKF methodology,
the covariance matrix $\mathbf{P}_t$ is propagated via the Riccati equation associated with the linearized error model in Eq.~\eqref{equ: log-linear eqn}:
$
    \dot{\mathbf{P}}_t = \mathbf{A}_t\mathbf{P}_t+\mathbf{P}\mathbf{A}_t^T+\mathbf{\bar{Q}}_t
$,
where
$
    \mathbf{\bar{Q}}_t \triangleq
    \begin{bmatrix}
        \mathbf{Ad}_{\mathbf{\bar{X}}_t} & \mathbf{0}_{12\times 6}
        \\
        \mathbf{0}_{6\times 12} &\mathbf{I}_6
    \end{bmatrix}
    \text{Cov}(\mathbf{w}_t)
    \begin{bmatrix}
        \mathbf{Ad}_{\mathbf{\bar{X}}_t} & \mathbf{0}_{12\times 6}
        \\
        \mathbf{0}_{6\times 12} &\mathbf{I}_6
    \end{bmatrix}^T.
$

\noindent {\it \textbf{Remark 1}} \textbf{(Group affine property):}
Without IMU biases, the continuous process model in Eq.~\eqref{equ: process model} is group affine as defined in~\cite{7523335}.
Thus, without biases and in the deterministic case,
{the linear error dynamics in Eq.~\eqref{equ: log-linear eqn} is exact and independent of the true state, and the covariance propagation is exact.
Such features are different from the standard EKF whose linearization accuracy relies on estimation error.}

\subsection{Continuous-Phase Measurement Models and Update Step}

This subsection formulates the two measurements in Eqs.~\eqref{equ: position measuremment} and \eqref{equ:haha} into the right-invariant observation form defined in~\cite{7523335} and introduces the update step of the proposed InEKF at time $t_n$.
{These treatments result in an error update equation that is independent of the true state.}

\subsubsection{Right-invariant orientation based measurement}
The orientation based measurement in Eq.~\eqref{equ:haha} can be rewritten into the following right-invariant observation form:
\begin{equation}
\label{equ: right inv foot orientation}
\begin{aligned}
    \underbrace{\begin{bmatrix}
    \mathbf{h}_R(\mathbf{\tilde{q}})_{t_n}
    \scriptsize{
    \begin{bmatrix}
    0
    \\
    0
    \\
    1
    \end{bmatrix}
    }
    \\
    \mathbf{0}_{3 \times 1}
    \end{bmatrix}}_{\mathbf{Y}_{1,t_n}} =
    \mathbf{X}_{t_n}^{-1}
    \underbrace{\begin{bmatrix}
    \mathbf{\tilde{R}}^{DRS}_{t_n}
    \scriptsize{
    \begin{bmatrix}
    0
    \\
    0
    \\
    1
    \end{bmatrix}
    }
    \\
    \mathbf{0}_{3 \times 1}
    \end{bmatrix}
    }_{\mathbf{d}_{1,t_n}}
    +
    \begin{bmatrix}
        \mathbf{V}_{1,t_n} 
        \\
        \mathbf{0}_{3 \times 1}
    \end{bmatrix}
\end{aligned}
\end{equation}
with
$
\mathbf{V}_{1,t_n} = 
\mathbf{{R}}_{t_n}^T
        (\mathbf{\tilde{R}}^{DRS}_{t_n}
    \begin{bmatrix}
    0
    ~
    0
    ~
    1
    \end{bmatrix}^T)_\times
    \mathbf{w}_{t_n}^{DRS}
    +
     \tfrac{\partial \mathbf{h}_{R,3}}{\partial \mathbf{q}}(\tilde{\mathbf{q}}_{t_n})
     \mathbf{w}_{t_n}^{q}
$.

\subsubsection{Right-invariant position measurement}
The position measurement in Eq.~\eqref{equ: position measuremment} can be expressed as~\cite{hartley2020contact}:
\begin{equation}
\label{equ: sim update}
\begin{aligned}
    \underbrace{\begin{bmatrix}
    {\mathbf{h}_p}(\Tilde{\mathbf{q}}_{t_n})
    \\
    0
    \\
    1
    \\
    -1
    \end{bmatrix}}_{\mathbf{Y}_{2,t_n}}
    = \mathbf{X}^{-1}_{t_n}
    \underbrace{\begin{bmatrix}
    \mathbf{0}_{3\times1}
    \\
    0
    \\
    1
    \\
    -1
    \end{bmatrix}}_{\mathbf{d}_{2,t_n}}
    +
    \begin{bmatrix}
    \frac{\partial \mathbf{h}_p}{\partial \Tilde{\mathbf{q}}} (\tilde{\mathbf{{q}}}_{t_n})
    \mathbf{w}^q_{t_n}
    \\
    \mathbf{0}_{3\times 1}
    \end{bmatrix}.
\end{aligned}
\end{equation}

\subsubsection{Update}
At time $t_n$,
the updated estimates and covariance, denoted as ($\mathbf{\bar{X}}_{t_n}^\dagger, \boldsymbol{\bar{\theta}}_{t_n}^\dagger)$ and $ \mathbf{P}_{t_n}^\dagger$, are given by~\cite{7523335}:
\begin{equation}
\small
\begin{gathered}
    \mathbf{\bar{X}}_{t_n}^\dagger 
    = \text{exp}
    \left(
    \mathbf{L}_{t_n}^\xi
    \mathbf{z}_{t_n}
    \right)
    \mathbf{\bar{X}}_{t_n},
   ~
    \boldsymbol{\bar{\theta}}_{t_n}^\dagger = \boldsymbol{\bar{\theta}}_{t_n}+\mathbf{L}_{t_n}^{\zeta}
    \mathbf{z}_{t_n}
    ,
    ~
\mathbf{P}_{t_n}^\dagger = (\mathbf{I} - \mathbf{L}_{t_n} \mathbf{H}_{t_n}) \mathbf{P}_{t_n},
\end{gathered}
\label{equ: state update}
\end{equation}
where
$\mathbf{L}_{t_n}
\triangleq
\begin{bmatrix}
(\mathbf{L}_{t_n}^\xi)^T,
(\mathbf{L}_{t_n}^\zeta)^T
\end{bmatrix}^T$
is filter gain,
$\mathbf{H}_{t_n}$
is the observation matrix,
and
$
\mathbf{z}_{t_n}
\triangleq
\begin{bmatrix}
(\mathbf{\bar{X}}_{t_n} \mathbf{Y}_{1,t_n} - \mathbf{d}_{1,t_n})^T,~
(\mathbf{\bar{X}}_{t_n} \mathbf{Y}_{2,t_n} - \mathbf{d}_{2,t_n})^T
\end{bmatrix}^T$.

To derive the observation matrix $\mathbf{H}_{t_n}$,
we first decompose it into
$
    \mathbf{H}_{t_n}
    =
    \begin{bmatrix}
    \mathbf{H}_{1,t_n}^T,
    ~
    \mathbf{H}_{2,t_n}^T
    \end{bmatrix}^T
$,
where 
$\mathbf{H}_{1,t_n} \in \mathbb{R}^{6 \times 12}$ and $\mathbf{H}_{2,t_n}\in \mathbb{R}^{6 \times 12}$ are respectively associated with the measurement models in \eqref{equ: right inv foot orientation} and \eqref{equ: sim update}.
Since the measurement models are not explicitly dependent on biases, the matrix $\mathbf{H}_{i,t_n}$ ($i=1,2$) can be further decomposed as
$
\mathbf{H}_{i,t_n}
\triangleq
\begin{bmatrix}
        \tilde{\mathbf{H}}_{i,t_n},
        ~
      \mathbf{0}_{3\times 6}~
      ;
      \mathbf{0}_{3\times 12},
        ~
      \mathbf{0}_{3\times 6}
\end{bmatrix}
$,
where the element $\mathbf{0}_{3\times 6}$ correspond to the bias terms and the element $\mathbf{0}_{3\times 12}$ could be removed if {a reduced-dimensional filter gain is instead used as in~\cite{hartley2020contact}}.
Based on the right-InEKF methodology~\cite{7523335},
we obtain the submatrix 
$
\tilde{\mathbf{H}}_{i,t_n}
$ via
$
\begin{aligned}
\tilde{\mathbf{H}}_{i,t_n}\boldsymbol{\xi}_{t_n}=-(\boldsymbol{\xi}_{t_n})^\wedge \mathbf{d}_{i,t_n}
\end{aligned}$:
$ 
\tilde{\mathbf{H}}_{1,t_n} \triangleq [(\mathbf{R}^{DRS}_{t_n}~[0, 0, 1]^T)_{\times},~\mathbf{0}_{3 \times 9}]$
and
$\tilde{\mathbf{H}}_{2,t_n} \triangleq [\mathbf{0}_{3\times 6},~ -\mathbf{I}_{3},~ \mathbf{I}_{3}]$.

To compute $\mathbf{L}_{t_n}$,
the linearized error update equation is obtained based on the update equation (Eq.~\eqref{equ: state update}) as:
\begin{equation}
\small
    \begin{bmatrix}
    \boldsymbol{\xi}_{t_n}^\dagger
    \\
    \boldsymbol{\zeta}_{t_n}^\dagger
    \end{bmatrix}
    =
    (\mathbf{I} {-} \mathbf{L}_{t_n} \mathbf{H}_{t_n})
    \begin{bmatrix}
    \boldsymbol{\xi}_{t_n}
    \\
    \boldsymbol{\zeta}_{t_n}
    \end{bmatrix}
    +
    {\mathbf{L}_{t_n}
    \begin{bmatrix}
        \mathbf{\bar{R}}_{t_n}
        \frac{\partial \mathbf{h}_{R,3}}{\partial {\mathbf{q}}_{t}} (\tilde{\mathbf{{q}}}_{t_n})
        \\
        \mathbf{0}_{3\times 1}
    \\
        \mathbf{\bar{R}}_{t_n}
        \frac{\partial \mathbf{h}_p}{\partial {\mathbf{q}}_{t}} 
        (\Tilde{\mathbf{{q}}}_{t_n})
        \\
        \mathbf{0}_{3\times 1}
    \end{bmatrix}
    \mathbf{w}^{q}_{t_n}},
    \label{equ: error update 1}
\end{equation}
with derivation given in Sec. 4 of supplementary material.

Then, applying the standard Kalman filtering methodology to this linear error update equation, we obtain the filter gain:
$
\begin{aligned}
\mathbf{L}_{t_n}
=\mathbf{P}_{t_n}\mathbf{H}_{t_n}^T\mathbf{S}_{t_n}^{-1}
\end{aligned},$
where
$
\mathbf{S}_{t_n} = \mathbf{H}_{t_n}\mathbf{P}_{t_n}\mathbf{H}_{t_n}^T+\mathbf{\bar{N}}_{t_n}$
,
$
\mathbf{\bar{N}}_{t_n} 
\triangleq
\text{diag}(\mathbf{\bar{N}}_{1,t_n},\mathbf{\bar{N}}_{2,t_n})
$,
$
\mathbf{\bar{N}}_{1,t_n} 
\triangleq
\mathbf{\bar{R}}_{t_n}
\tfrac{\partial \mathbf{h}_{R,3}}{\partial \mathbf{q}_{t}} (\mathbf{\Tilde{q}}_{t_n})
\text{Cov}(\mathbf{w}_{t_n}^q)
(\tfrac{\partial \mathbf{h}_{R,3}}{\partial \mathbf{q}_{t}} (\mathbf{\Tilde{q}}_{t_n}))^T
\mathbf{\bar{R}}_{t_n}^T
$,
and
$
\mathbf{\bar{N}}_{2,t_n} 
\triangleq
\mathbf{\bar{R}}_{t_n}
\tfrac{\partial \mathbf{h}_p}{\partial \mathbf{q}_{t}} (\mathbf{\Tilde{q}}_{t_n})
\text{Cov}(\mathbf{w}_{t_n}^q)
(\tfrac{\partial \mathbf{h}_p}{\partial \mathbf{q}_{t}} (\mathbf{\Tilde{q}}_{t_n}))^T
\mathbf{\bar{R}}_{t_n}^T$.

\noindent {\it \textbf{{Remark 2}}} \textbf{(Independence of true state):}
The linearized error update equation (Eq.~\eqref{equ: error update 1})
is independent of the true state $\mathbf{X}_t$ and $\boldsymbol{\theta}_t$ in the deterministic case.
This is because both measurement models satisfy the right-invariant observation form with respect to $\mathbf{X}_t$
and are independent of $\boldsymbol{\theta}_t$ and because the update equation of $\bar{\mathbf{X}}_t$ is in the exponential form as prescribed by the InEKF methodology~\cite{7523335}.

\subsection{Discrete Process Model and Propagation Step}
Without loss of generality and for simplicity, suppose that the foot-landing events and the updates do not coincide.
Thus,
the proposed filtering for the state jump focuses on estimate and covariance propagation without update.
Except for the true contact point position $\mathbf{p}^c_{t}$, the rest of the true state is continuous across a foot landing, as explained in Sec.~\ref{section: problem formulation}.

\subsubsection{Process model}

From the proposed jump dynamics in Sec.~\ref{section: problem formulation}, the stochastic jump dynamics of $\mathbf{X}_t$ can be approximately expressed as:
\begin{equation}
\begin{aligned}
   {\mathbf{{X}}_{t^+}} 
    &=
   {\mathbf{{X}}_{t}} 
   \begin{bmatrix}
       \mathbf{I}_{3} 
       &
       [\mathbf{0}_{3 \times 1},~\mathbf{0}_{3 \times 1},~ \mathbf{h}_{c}(\tilde{\mathbf{q}}_{t})]
       \\
       \mathbf{0}_{3 \times 3} &  \mathbf{I}_3
   \end{bmatrix}
   -
    {
   {\mathbf{{X}}_{t}}
   \begin{bmatrix}
        \frac{\partial \mathbf{h}_c}{\partial \mathbf{q}} (\mathbf{\Tilde{q}}_{t})
       \mathbf{w}_t^{q}
       \\
       \mathbf{0}_{3 \times 1}
   \end{bmatrix}^\wedge}
   \\
      & \triangleq
   \boldsymbol{\Delta}_{u_t}({\mathbf{{X}}_{t}})
   - 
    {
   {\mathbf{{X}}_{t}} (\mathbf{w}^{\Delta}_t)^\wedge},
 \end{aligned}
 \label{equ: jump map X}
\end{equation}
where the function $\boldsymbol{\Delta}_{u_t}({\mathbf{{X}}_{t}})$ is the deterministic jump dynamics, and
the encoder data $\mathbf{\Tilde{q}}_t$ serves as the input. 
As the biases are continuous under a jump event, $ {\boldsymbol{\theta}}_{t^+} = {\boldsymbol{\theta}}_t$ holds.

\noindent {\it \textbf{Remark 3}} \textbf{(Group affine property):}
The jump map $\boldsymbol{\Delta}_{u_t}$ of the state $\mathbf{X}_t$ possesses the discrete-time group affine property defined in~\cite{barrau2018invariant}, and is independent of IMU biases $\boldsymbol{\theta}_t$.
Thus, the jump dynamics of the error $\boldsymbol{\xi}_t$ is independent of the true state and is exactly linear.
Moreover, from the expression of $\boldsymbol{\Delta}_{u_t}$ in Eq.~\eqref{equ: jump map X}, we can see that $\boldsymbol{\Delta}_{u_t}$ is a group action on $SE_3(3)$,
under which the error =$\boldsymbol{\xi}_t$ naturally does not change.

\subsubsection{Error equation}
From Eq.~\eqref{equ: jump map X}, we obtain the dynamics of the logarithmic error $\boldsymbol{\xi}_t$ as:
$ \boldsymbol{\xi}_{t^+} = \boldsymbol{\xi}_{t} - \mathbf{Ad}_{\bar{\mathbf{X}}} \mathbf{w}^{\Delta}_t$.
Indeed, as analyzed in Remark 3, the error does not jump under $\boldsymbol{\Delta}_{u_t}$.
Also, $\boldsymbol{\zeta}_{t^+}=\boldsymbol{\zeta}_{t^-}$ holds since the IMU biases are continuous.

\subsubsection{Propagation}
Based on the deterministic portion of the jump model in Eq.~\eqref{equ: jump map X}, the propagation of the state estimate at a jump event is: $ \bar{\mathbf{{X}}}_{t^+} = \boldsymbol{\Delta}_{u_t}(\bar{\mathbf{{X}}}_{t})$ and $ \bar{\boldsymbol{\theta}}_{t^+} = \bar{\boldsymbol{\theta}}_t$.

With the linear error equation of $\boldsymbol{\xi}$ and $\boldsymbol{\zeta}$ across a jump,
the propagation of the covariance matrix is expressed as:
$
    \mathbf{\mathbf{P}}_{t^+} = \mathbf{\mathbf{P}}_{t}
+
\bar{\mathbf{Q}}^{\Delta}_t
$,
where
$
\bar{\mathbf{Q}}^{\Delta}_t
=
\begin{bmatrix}
    \mathbf{Ad}_{\bar{\mathbf{X}}} \text{Cov}(\mathbf{w}^{\Delta}_t) \mathbf{Ad}_{\bar{\mathbf{X}}}^T
    &
    \mathbf{0}_{3 \times 3}
    \\
    \mathbf{0}_{3 \times 3}
    &
    \mathbf{0}_{3 \times 3}
\end{bmatrix}
$.

The complete algorithm of the proposed right-InEKF is summarized as Algorithm 1 in supplementary material.

\noindent
{\noindent {\it \textbf{Remark 4}} \textbf{ (Imperfect InEKF):}}
In the presence of IMU biases, the proposed filter is no longer a ``perfect'' InEKF in the sense that the group affine and invariant form properties no longer hold for continuous phases.
Although the linear equation in Eq.~\eqref{equ: log-linear eqn} is no longer independent of the true state,
it depends on the true state only through the bias terms while the remaining part of the Jacobian matrix $\mathbf{A}_t$ is still {independent of the true state}.
Also, the measurement models are still independent of the true state $\mathbf{X}_t$ and $\boldsymbol{\theta}_t$
{as highlighted in Remark 2}.
For these reasons, {the linearization inaccuracy induced by the biases has a limited impact on the continuous-phase propagation and update.
Thus,} the ``imperfect InEKF'' with biases considered can still ensure rapid and accurate convergence under large errors, which is experimentally confirmed on DRS locomotion as reported in Sec.~\ref{section: experiments}.

\section{Observability and Convergence Analysis}
\label{section: observability analysis}

\subsection{Observability Analysis for Continuous Phases}
As measurement update is performed during continuous phases,
we only analyze the continuous-phase observability.

Recall that the deterministic continuous-phase dynamics in Eq.~\eqref{equ: process model} is group affine in the absence of IMU biases $\boldsymbol{\theta}_t$ ({Remark 1}).
Also, recall that the measurement models in Eqs.~\eqref{equ: right inv foot orientation} and \eqref{equ: sim update} are in the right-invariant observation form with respect to $\mathbf{X}_t$, regardless of the presence of biases ({Remark 2}).
Then, by Theorem 20 in~\cite{barrau2015non}, the observability of $\mathbf{X}_t$ for the complete continuous-phase system, which has both $\mathbf{X}_t$ and $\boldsymbol{\theta}_t$ as its state, is the same as that of the simplified continuous-phase system without IMU biases.
Thus, by Theorem 5 in~\cite{7523335}, {the local observability of $\mathbf{X}_t$} for the complete system can be determined by the couple ($\mathbf{A}$, $\mathbf{H}$), with $\mathbf{A}$ and $\mathbf{H}$ updated with bias-related terms removed
(see Sec. 6 of supplementary material).

With $\Delta t$ the duration of one propagation step, the discrete state transition matrix $\boldsymbol{\Phi}$ is given by $\boldsymbol{\Phi} = \text{expm}(\mathbf{A}_t \Delta t)$~\cite{7523335} (see Sec. 6 of supplementary material for the expression of $\boldsymbol{\Phi}$).
Then, from {$\mathfrak{O} = 
    \begin{bmatrix}
        (\mathbf{H})^T,
        ~
        (\mathbf{H}\boldsymbol{\Phi})^T,
        ~
        (\mathbf{H}\boldsymbol{\Phi}^2)^T,
        ~
        \hdots
    \end{bmatrix}^T$}, we have:
\begin{equation}
\begin{aligned}
    \mathfrak{O} &= 
    {
    \begin{bmatrix}
        (\mathbf{R}^{DRS}~[0, 0, 1]^T)_{\times}& \mathbf{0}_{3\times 3} & \mathbf{0}_{3\times 3} & \mathbf{0}_{3\times 3}
        \\
        \mathbf{0}_{3\times 3} & \mathbf{0}_{3\times 3} & -\mathbf{I}_3 & \mathbf{I}_3
        \\
        (\mathbf{R}^{DRS}~[0, 0, 1]^T)_{\times}& \mathbf{0}_{3\times 3} & \mathbf{0}_{3\times 3} & \mathbf{0}_{3\times 3}
        \\
        -\frac{1}{2} (\mathbf{g})_{\times} \Delta t^2 & -\mathbf{I}_3 \Delta t & -\mathbf{I}_3 & \mathbf{I}_3
        \\
        \hdots & \hdots & \hdots &\hdots
    \end{bmatrix}.
    }
\end{aligned}
    \label{Eqn: Obs matrix}
\end{equation}

As the first two columns of {$(\mathbf{g})_{\times}$} are linearly independent, the base roll and pitch angles are observable.
Because all columns in the second column block of {$\mathfrak{O}$} are linear independent, the base velocity $\mathbf{v}_t$ is observable.
Yet, as the last two column blocks are linearly dependant,
the base position $\mathbf{p}_t$ and contact point position $\mathbf{p}^c_t$ are unobservable.

The third column of $(\mathbf{g})_{\times}$ is always zero because only its $z$-component is nonzero.
Then, if the surface is non-horizontal, (i.e., the third column of $(\mathbf{R}^{DRS}~[0, 0, 1]^T)_{\times}$ is not all zero), the yaw will be observable; 
otherwise, it is unobservable.

From the expression of $\mathfrak{O}$, we also know that: a) the contact velocity $\mathbf{v}^c_t$ does not affect observability; b) either measurement model ensures observable base roll and pitch; c)
the proposed measurement in Eq.~\eqref{equ: right inv foot orientation} renders base yaw observable; and d) the previous measurement in Eq.~\eqref{equ: sim update} makes base velocity observable.

\subsection{Convergence Property for Hybrid Error System}
\label{section: convergece analysis}

The proposed convergence analysis for the hybrid error system is built upon previous analysis of the InEKF as a deterministic observer for systems without state-triggered jumps~\cite{7523335}.
Different from the previous work, this subsection analyzes the effects of the jumps on the error convergence for the overall hybrid error system.

We first analyze the error evolution across the deterministic discrete jump of the system.
Analyzing the state evolution across discrete, state-triggered jumps (e.g., foot-landing impacts) is typically complex~\cite{iqbal2020provably}.
Yet, since the jump map $\boldsymbol{\Delta}_{u_t}$ is a group action, the error $ \boldsymbol{\xi}_t$ does not jump under $\boldsymbol{\Delta}_{u_t}$ despite the jump of the true state $\mathbf{X}_t$. 
Also, the bias error $\boldsymbol{\zeta}_t$ is continuous across a jump event. 
Thus, the hybrid, deterministic error dynamics is essentially continuous for all time, and its error convergence is equivalent to that of the deterministic continuous phases.

For continuous phases, the proposed filter meets the group affine condition and invariant observation form without biases, as discussed in {Sec.~\ref{section: InEKF}}.
Thus, by the theory of InEKF~\cite{7523335}, the proposed filter is {locally} asymptotically convergent for the observable variables of the deterministic continuous phases without biases.
Accordingly, the local asymptotic convergence of the hybrid, deterministic filter system is guaranteed in the absence of biases.

This analysis also supports the local asymptotic convergence of the existing InEKF~\cite{hartley2020contact} designed for static surface locomotion, because the jump model in~\cite{hartley2020contact} is a group action and its continuous-phase design also satisfies the group affine and invariant observation conditions without biases.

\section{EXPERIMENTS}
\label{section: experiments}

\subsection{Experimental setup}

\begin{figure}[t]
    \centering
    \includegraphics[width=0.81\linewidth]{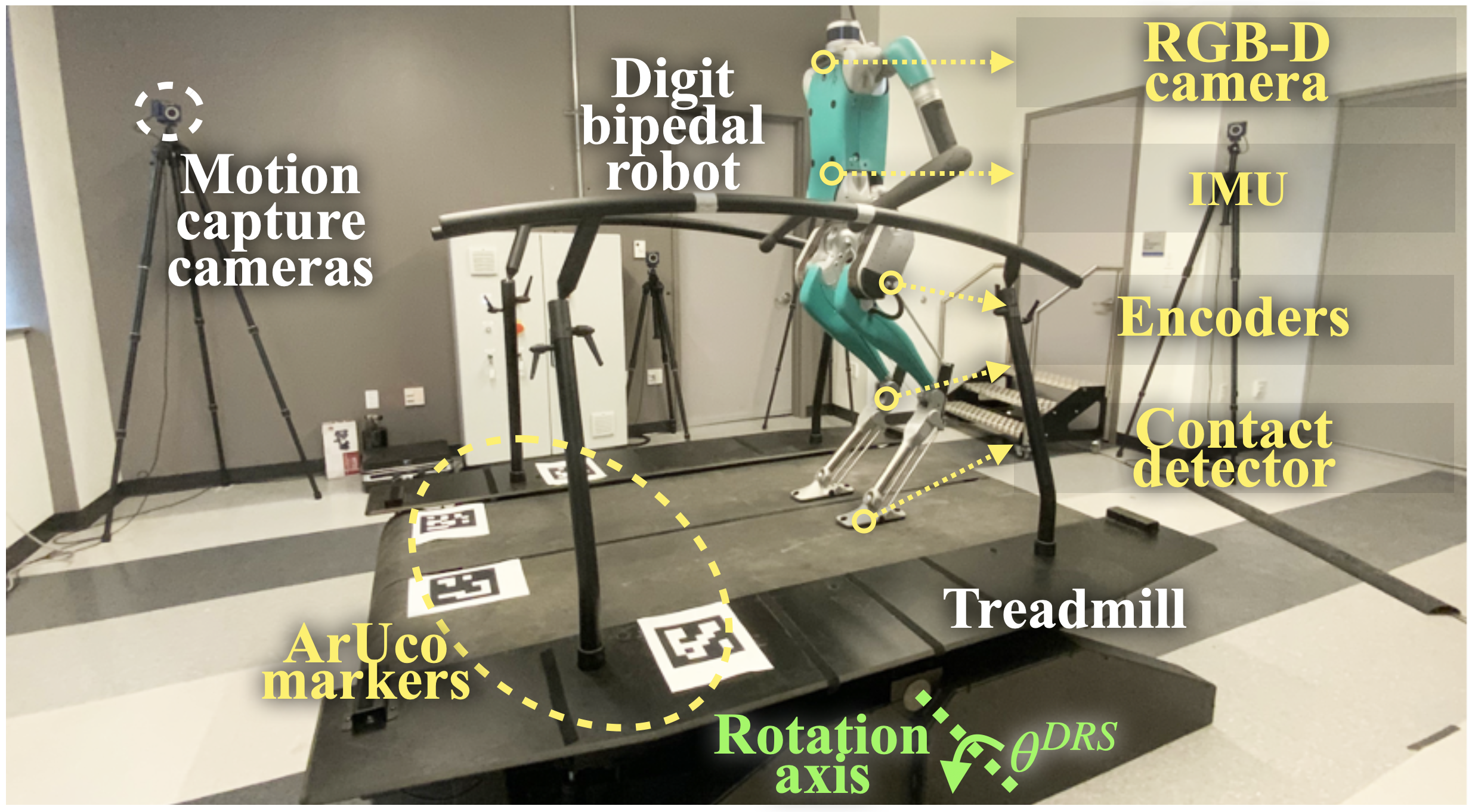}
    \caption{Experimental setup that includes a Digit bipedal humanoid robot and a pitching treadmill (i.e., DRS).}
    \label{Fig:experiment_setup}
\end{figure}

\begin{figure}[t]
    \centering
    \includegraphics[width=0.85\linewidth]{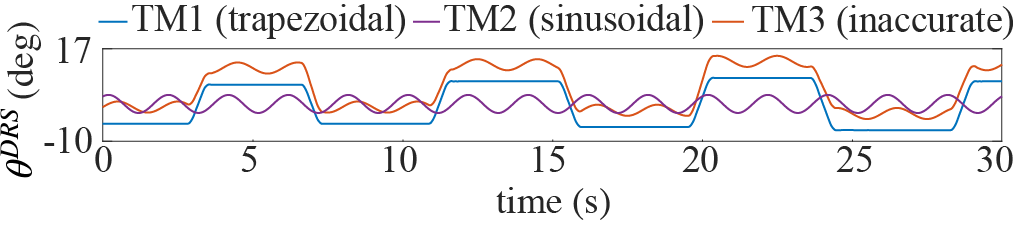}
    \caption{Profiles (TM1)-(TM3) of the treadmill pitch angle $\theta^{DRS}(t)$.}
    \label{Fig:v5_treadmill_motion}
\end{figure}

The setup for experimental data collection (Fig.~\ref{Fig:experiment_setup}) is:

\noindent \textbf{Treadmill (i.e, the tested DRS).}
A split-belt Motek M-gait treadmill is used as a DRS.
Its dimension is 2.3 m$\times$1.82 m$\times$0.5 m.
To emulate a rocking ship in sea waves, it performs a whole-body pitching motion without belt translation.

\noindent \textbf{Robot.}
The Digit robot is $1.6$ m tall, 
and each leg's kinematic chain used by the filter has $12$ joints.
Different robot movements are tested: \textbf{(RM1)} stepping and \textbf{(RM2)} standing.
The robot is about 0.8 m away from the treadmill center.

\noindent \textbf{Treadmill motion profiles.}
To test filter performance under different DRS motions,
two different profiles of the treadmill's pitch angle $\theta^{DRS}$ (Fig. 3) are tested: \textbf{(TM1)} a non-periodic trapezoidal wave, $f_{trap}(t)$,
and
\textbf{(TM2)} a sine wave $2.5^\circ \sin(\pi t)$.
Under {(TM1)} and {(TM2)}, the maximum contact point speeds $\| \mathbf{v}^c_t \|$ are respectively $0.41$ m/s and $0.11$ m/s.
To test the filter's robustness under surface motion inaccuracy, a fictitious profile is considered:
\textbf{(TM3)} $\theta^{DRS}(t) =f_{trap}(t) +5.1^\circ + 1.7^\circ \sin(\pi t)$
with the actual profile (TM1) used in experiments.
Figure~\ref{Fig:v5_treadmill_motion} shows all profiles.

\begin{figure}[t]
    \centering
    \includegraphics[width=0.85\linewidth]{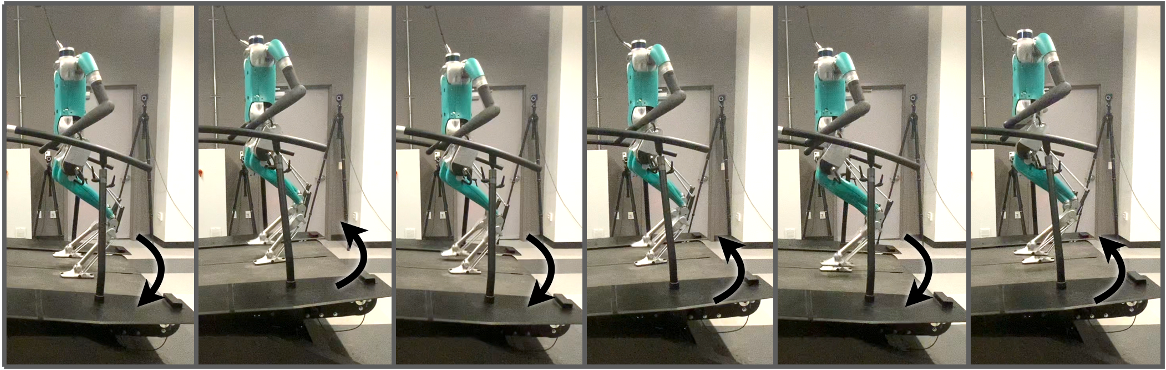}
    \caption{Time-lapse figures of Digit walking on a rocking treadmill.
    {The black arrow indicates the treadmill's direction of rotation}.}
    \label{Fig:digit_time_elapse}
\end{figure}

\begin{figure}[t]
    \centering
    \includegraphics[width=1\linewidth]{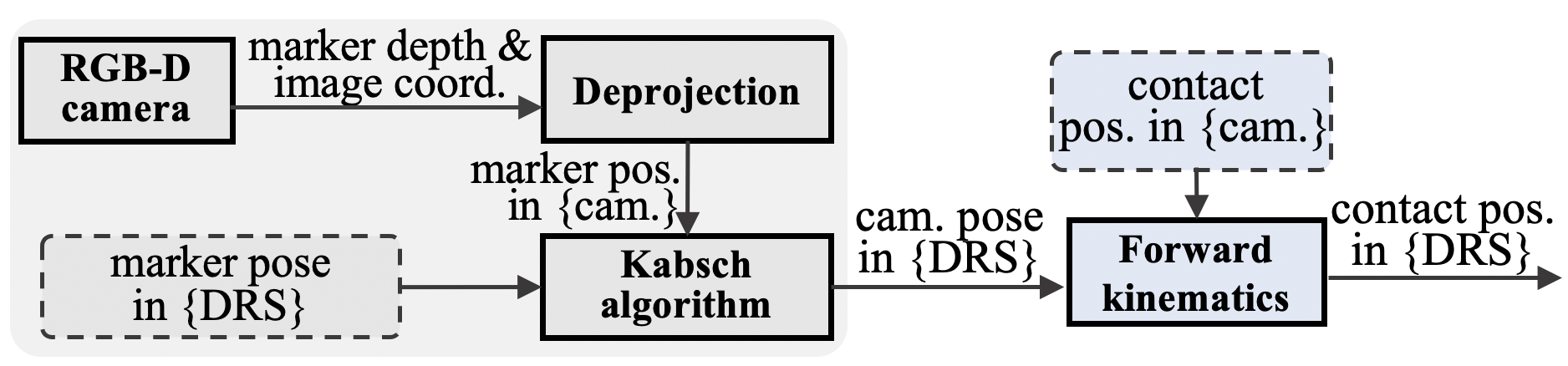}
    \caption{Procedure of obtaining the 3-D contact point position in the DRS frame using the ArUco markers and the robot's on-board RGB-D camera.}
    \label{Fig:flowchart_to_compute_vd}
\end{figure}

\begin{figure}[t]
    \centering
    \includegraphics[width=0.95\linewidth]{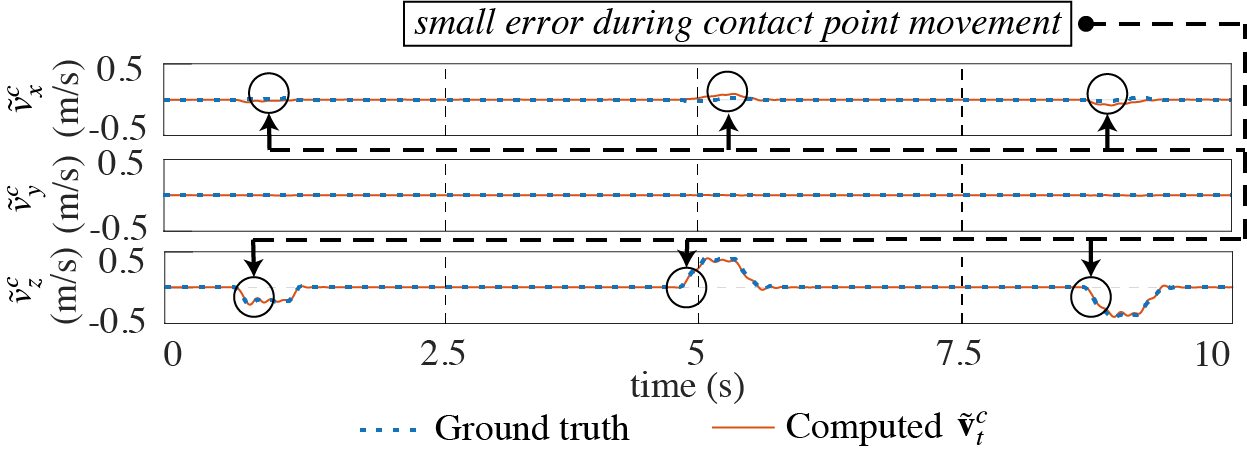}
    \caption{Validation results of the proposed method for obtaining the contact point velocity $\mathbf{\tilde{v}}_t^c \triangleq [\tilde{v}^c_x,\tilde{v}^c_y,\tilde{v}^c_z]^T$ under Case C.
    The velocity along the y-direction, $\tilde{v}^c_y$, is zero because the treadmill does not move in that direction.}
    \label{Fig:v2_foot_vel}
\end{figure}

\begin{table}[t]
    \centering
    \caption{Noise standard deviation for inekf-srs and inekf-drs.}
\begin{tabular}{ p{3.7cm}|p{1.7cm}|p{1.8cm}  }
\hline
\hline
\centering
Measurement type & InEKF-SRS & InEKF-DRS\\
\hline
\centering Linear acc. (m/s$^2$) & \centering  $0.4$  & ~~~~~~~$0.4$ \\
\centering Angular vel. (rad/s) & \centering $0.01$    & ~~~~~~$0.01$ \\
\centering Acc. bias (m/s$^3$) & \centering $0.001$ & ~~~~~$0.001$ \\
\centering Gyroscope bias (rad/s$^2$) & \centering  $0.0001$  & ~~~~$0.0001$ \\
\centering Contact vel. (m/s) &\centering $0.01$  & ~~~~~~$0.01$ \\
\centering Encoder $(^\circ)$ &\centering  $1$ & ~~~~~~~~$1$ \\
\centering DRS orientation {($^\circ$)} & \centering N/A &~~~~~~~~$1$
\\
\hline
\end{tabular}
\label{Table: covariance}
\end{table}

\noindent \textbf{On-board sensors used.}
Digit's on-board sensors used (Fig. 3) are: an IMU, joint encoders, a RealSense RGB-D camera, and the robot's {proprietary} contact detector.
The camera returns data at 15 Hz, and the remaining sensors stream data at the same rate within $60$-$90$ Hz.
Cortex motion capture cameras provide the ground truth.
ArUco markers are attached to the treadmill, emulating the real-world scenario where legged robots that navigate within a DRS (e.g., a vessel at sea) can only see landmarks attached to the DRS but not any landmarks on the earth's ground.
The markers are sensed by the camera to obtain the camera pose in the treadmill frame, which is then used to compute contact point velocity as explained in the next subsection.

\noindent \textbf{Data collection cases.}
Figure~\ref{Fig:digit_time_elapse} shows screenshots of experiments.
The filter is simulated in MATLAB using four experimentally collected data sets under different robot and treadmill motions:
Case A: Combination of (RM1) and (TM1);
Case B: Combination of (RM1) and (TM2);
Case C: Combination of (RM2) and (TM1); and
Case D: Combination of (RM1) and (TM3), where 
the actual profile is (TM1) but the filter uses the inaccurate data (TM3).
The experiment video is available at: 
\href{https://youtu.be/ScQIBFUSKzo}{https://youtu.be/ScQIBFUSKzo}.

\subsection{Filter Setting}

\noindent \textbf{Filters compared.}
The proposed filter (denoted as ``InEKF-DRS'') is compared with an InEKF designed for locomotion on a static rigid surface~\cite{hartley2020contact} (denoted as ``InEKF-SRS'').
The InEKF-SRS models the deterministic contact point motion as $ \dot{\mathbf{p}}^c=\mathbf{0}$, and uses the position measurement in Eq.~\eqref{equ: sim update} alone.
It renders the base orientation (except for yaw) and velocity observable.
It has realized substantially faster convergence under large errors during stationary surface locomotion, as compared with EKF-based method~\cite{bloesch2013state}.
Also, the proposed InEKF-DRS is compared with an EKF-based filter, which we formulate by augmenting the existing EKF designed for static surfaces~\cite{bloesch2013state} to explicitly handle nonstationary surfaces. Details of the augmentation and comparison results are in Sec. 8.3 of supplementary material. The rest of this section focuses on comparing InEKF-DRS and InEKF-SRS.

\noindent \textbf{Contact point velocity computation.}
The contact point velocity $\tilde{\mathbf{v}}^c$ serves as an input to the continuous-phase process model of the proposed InEKF-DRS.
To obtain the contact point velocity $\tilde{\mathbf{v}}^c$ (see Fig.~\ref{Fig:flowchart_to_compute_vd}),
we first obtain the camera pose in the DRS frame by processing the features of the ArUco markers in the camera images, which we then use to compute the 3-D contact point position in the DRS frame ($^{DRS}\mathbf{p}^{c}$) through forward kinematics.
Next, we estimate the contact point velocity $\mathbf{\tilde{v}}^c$ based on Eq.~\eqref{eqn: vd} using the known treadmill motion data.
Details of this procedure are in Sec. 7 of supplementary material.
Results in Fig.~\ref{Fig:v2_foot_vel} validate the accuracy of the proposed contact point velocity sensing.

\noindent \textbf{Covariance settings.}
Table~\ref{Table: covariance} shows the noise standard deviation (SD) of both filters.
The SD for the accelerometer, gyroscope, and their corresponding biases are obtained from the manufacturer's manual with a slight adjustment for better performance. 
The SD for the encoder readings is adopted from the previous filter~\cite{hartley2020contact} designed for a similar robot.
The SD for the contact-point velocity and orientation-based measurement are tuned for a reasonable performance.
The initial value of the covariance $\mathbf{P}$ is set as an identity matrix.

\noindent \textbf{Initial estimation errors.}
For a fair comparison, the two filters are simulated under the same large range of initial estimation errors.
The initial velocity and orientation errors in each direction are respectively uniformly distributed within [$- 1.5$,$1.5$] m/s and [$-1$,$1$] rad.

\begin{figure*}[t]
    \centering
    \includegraphics[width=1\linewidth]{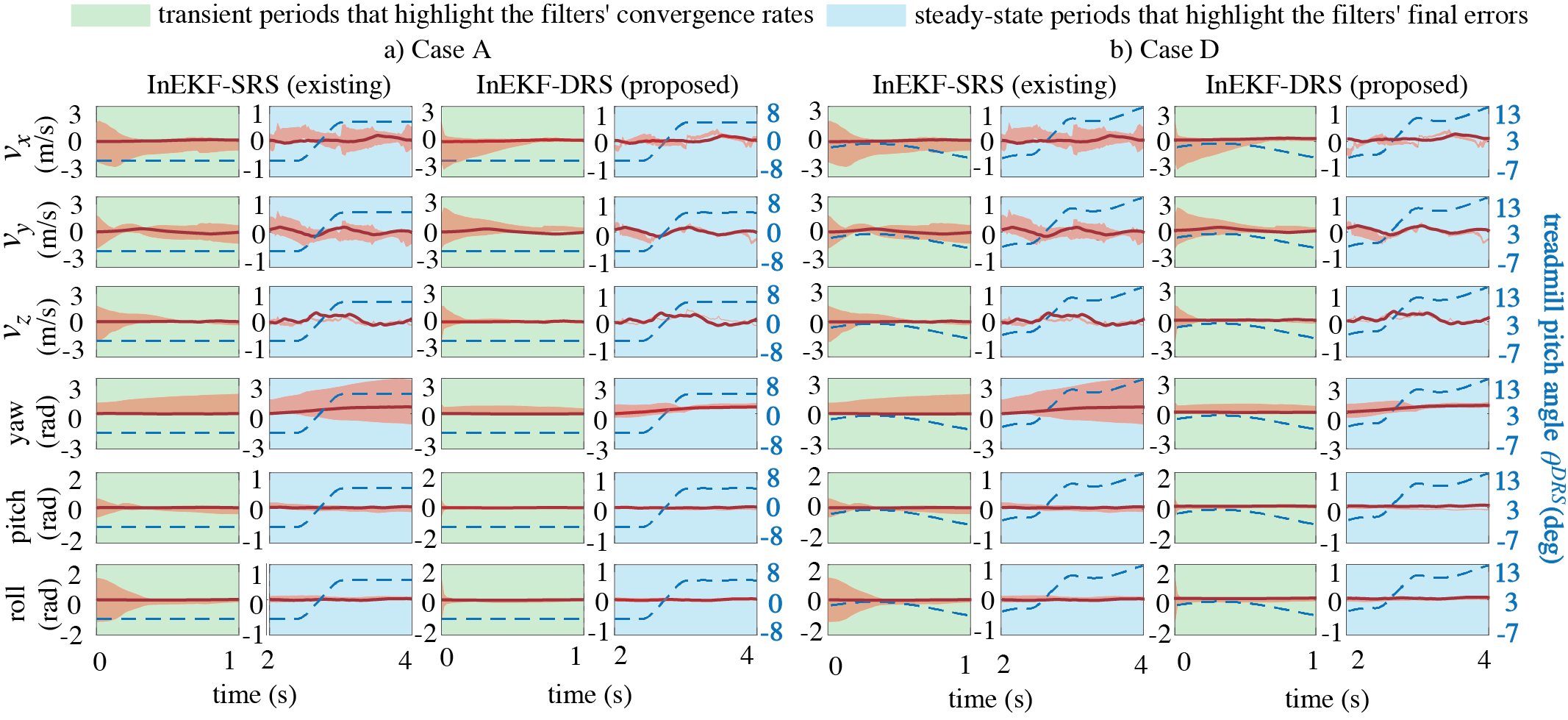}
    \caption{Base velocity and orientation estimation results of the two filters, InEKF-DRS and InEKF-SRS, for Cases A and D.
    The red, shaded area indicates the range of the state estimates for 10 runs.
    The red, solid line is the ground truth.
    The blue, dashed line is the treadmill angle.}
    \label{fig: CasesA-D}
\end{figure*}

\subsection{Computation Time Comparison}

In MATLAB, both filters take less than $1$ ms to compute one estimation cycle (i.e., one propagation and one update step),
confirming their validity for real-time estimation.

\subsection{Convergence Rate and Yaw Observability Comparison}
Figure~\ref{fig: CasesA-D}-a) displays the estimation results of InEKF-DRS (proposed) and InEKF-SRS under Case A where the treadmill stays at a pitch angle of $-8^{\circ}$ for approximately 2.8 sec and then begins to pitch until reaching $+8^{\circ}$ in 0.5 sec.

Both filters drive the error of base roll, pitch, and velocity {closer to zero}, indicating their observability as predicted in Sec.~\ref{section: observability analysis} and previous work~\cite{bloesch2013state,hartley2020contact}.
In terms of the convergence rates for these variables, subplot a) shows that the proposed InEKF-DRS is {faster than InEKF-SRS}, driving the error close to zero within 1 sec.
This is because InEKF-DRS considers the surface motion and has an additional measurement (Eq.~\eqref{equ: right inv foot orientation}) that corrects estimates.

Under InEKF-DRS, the yaw estimate converges close to the ground truth in approximately 3 sec, which supports the observability analysis in Sec.~\ref{section: observability analysis} that the yaw angle is observable if the DRS/treadmill is not horizontal.
Yet, the yaw convergence is slower than pitch and roll, possibly because both observations in Eqs.~\eqref{equ: right inv foot orientation} and \eqref{equ: sim update} help correct the roll and pitch estimates whereas only the former corrects the yaw estimate.
Finally, as previously revealed~\cite{bloesch2013state}, the yaw error divergence under InEKF-SRS confirms that the base yaw is indeed non-observable with InEKF-SRS.

\begin{figure}[t]
    \centering
    \includegraphics[width=0.87\linewidth]{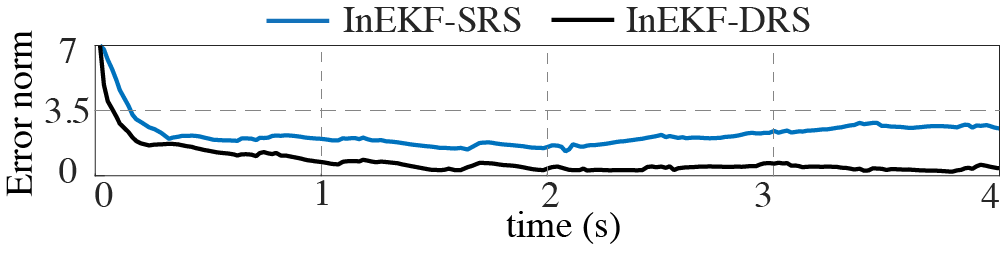}
    \caption{Accuracy comparison of InEKF-SRS and InEKF-DRS (proposed) for the estimation of base velocity and roll and pitch angles under Case A.
    }
    \label{Fig:error_norm_v5}
\end{figure}

\subsection{Accuracy Comparison}

Table~\ref{Table: RMSE} shows the comparison of the root-mean-square (RMS) estimation errors for base orientation (including yaw) and velocity under Case A.
Figure~\ref{Fig:error_norm_v5} shows the corresponding time evolution of the errors for base roll, pitch, and velocity under Case A.
The table and the figure show that the proposed InEKF-DRS is more accurate in velocity and orientation estimation compared with InEKF-SRS.

\subsection{Performance under Different DRS and Robot Movements}
\label{subsection:Filter Performance under Different Surface Movements and Robot Movements}

Figures 1-a) and 1-b) in supplementary material respectively show the estimation results of the two filters under Case  B (where the treadmill motion is different from Case A) and Case C (where the robot stands on the treadmill instead of walking as in Case A).
The plots show that the performance comparison of the two filters under Cases B and C are similar to Case A (i.e., Fig.~\ref{fig: CasesA-D}-a)), in terms of convergence rate and accuracy, indicating the InEKF-DRS can effectively handle different DRS and robot movements.

Comparing the yaw estimate under the InEKF-DRS in Cases A-C, we notice that the yaw estimate in Case C converges faster than Cases A and B.
In Case C, the treadmill remains horizontal for the first 10 sec, during which the yaw estimate does not converge. 
Yet, once the treadmill begins to rock at $t=10$ sec, the yaw estimate converges close to the ground truth within 1 sec, 
whereas it takes about 3 sec for the yaw estimate to enter a similar neighborhood under Cases A and B.
This might be due to the fact that in Case C, by the time the treadmill begins to pitch, the estimates of the rest observable state are already sufficiently accurate, making the yaw error correction faster than Cases A and B.

\begin{table}[t]
\centering
\caption{RMS error comparison under Case A.} 
\begin{tabular}{ p{2.5cm}|p{1.8cm}|p{1.8cm}  }
\hline
\hline
\centering
State variables & {\centering InEKF-SRS}  & {\centering  InEKF-DRS} \\
\hline
\centering $v_x$ (m/s)  & \centering $0.3320$ &  ~~~{\centering{$0.2051$}} \\
\centering $v_y$ (m/s) & \centering $0.2488$   &  ~~~{\centering $0.1955$} \\
\centering $v_z$ (m/s)  &\centering $0.1438$ & ~~~{\centering $0.1025$} \\
\centering yaw (rad) & \centering $0.9294$ & ~~~{\centering $0.2516$ }\\
\centering pitch (rad) &\centering $0.0897$  & ~~~{\centering $0.0413$} \\
\centering roll (rad) & \centering $0.1365$ & ~~~{\centering $0.0318$ } \\
\hline
\end{tabular}
\label{Table: RMSE}
\end{table}

\subsection{Robustness Assessment}
\label{subsection: Robustness Assessment}
Results from Cases A (Fig.~\ref{fig: CasesA-D} a) and D (Fig.~\ref{fig: CasesA-D} b) confirm the robustness of the proposed InEKF-DRS under inaccurate surface pose knowledge.
Case D emulates the scenario where the DRSes motion monitoring system fails to provide accurate DRS pose.
Subplots a and b show that the filter performance (e.g., convergence rate, accuracy, and yaw observability) under Case D is similar to that under Case A.
Specifically, the velocity estimate under InEKF-DRS converges to the ground truth in all directions within 1 sec.
Also, the orientation convergence rates are similar: the roll and pitch estimates converge close to the ground truth within 0.3 sec, and the yaw angle converges within 3 sec.
Longer periods (10-30 sec) of estimation results for Cases A-D are shown and discussed in supplementary material.

\section{DISCUSSION}
\label{section: discussion}

This study has designed an InEKF that estimates the orientation and velocity of a bipedal robot that walks on a DRS with a known, relatively significant motion, by fusing the known surface pose and the leg, visual, and inertial odometries.
Similar to the InEKF~\cite{hartley2020contact} and EKF~\cite{bloesch2013state} designed for stationary surfaces, the filter uses the IMU motion dynamics as the process model, and the 3-D contact point position and leg kinematics to form a measurement model.
Different from the previous work, the proposed contact-point process model does not assume that the contact point is static, but instead explicitly considers its movement in the world.
Also, we have introduced a right-invariant measurement model based on the rotational kinematic relationship between the surface and support foot.
Thanks to these features, the filter ensures accurate estimation under relatively large surface motion and estimation errors, as shown by the RMS errors in Table II and the state trajectories in Figs. \ref{fig: CasesA-D} and \ref{Fig:error_norm_v5}.

The proposed filter is suitable for a DRS with a relatively accurately known surface pose profile, but may not be effective under overly inaccurate or unknown profiles.
One potential solution is to extend this filter to estimate the surface pose, by 
constructing a matrix Lie group that includes the surface pose in the state and formulating an InEKF with fundamental benefits.

This study also assumes the robot's feet do not persistently and significantly slip on the surface. 
When the surface is slippery~\cite{trkov2019inertial}, the support foot may move relative to the surface, causing discrepancy between the actual robot movement and the models.
Yet, the proposed method could be extended to address slippage during DRS locomotion by incorporating existing techniques~\cite{teng2021legged,kim2021legged} such as using an RGB-D sensor to measure the base velocity~\cite{teng2021legged}.

\section{CONCLUSION}
\label{section: conclusion}
This paper has introduced a right-invariant extended Kalman filter for bipedal humanoid walking on a moving surface. 
The filter design explicitly considered the known surface movement and hybrid robot behaviors while enjoying the fundamental benefits of satisfying the attractive group-affine condition and invariant observation form in the absence of IMU biases.
Observability analysis for the continuous locomotion phases showed that the robot's base velocity and roll and pitch angles are observable, and the base yaw angle becomes observable when the DRS is not horizontal.
Stability analysis proved the asymptotic error convergence of these observable states for the hybrid deterministic system.
Experimental results of humanoid walking on a pitching treadmill validated the enhanced accuracy and convergence rate of the proposed filter over existing work, in the presence of large estimation errors and moderate DRS movement.

\section*{Acknowledgment}
The authors would like to thank M. Ghaffari and A. Saccon for constructive feedback on theoretical derivation.

\bibliography{Reference1}
\bibliographystyle{ieeetr}

\begin{IEEEbiography}[{\includegraphics[width=1.0in,height=1.2in,clip]{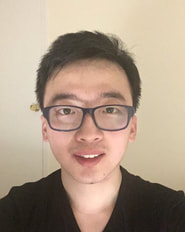}}]{Yuan Gao}
	received his B.S. degree in Mechanical Engineering from China Jiliang University, Hangzhou, China in 2014, and the M.S. degree in Mechanical Engineering from Arizona State University in 2016.
	He is currently a Ph.D. student in the Department of Mechanical Engineering at the University of Massachusetts Lowell, Lowell, MA, U.S.A..
	\end{IEEEbiography}

\begin{IEEEbiography}[{\includegraphics[width=1.0in,height=1.1in,clip]{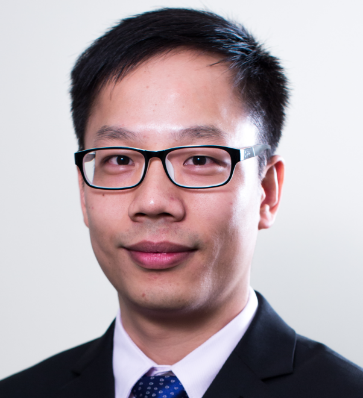}}]{Chengzhi Yuan}
received the B.S. and M.S. degrees from the South China University of Technology, Guangzhou, China, in 2009 and 2012, respectively, and the Ph.D. degree in mechanical engineering from
North Carolina State University, Raleigh, NC, U.S.A., in 2016.
He is currently an Assistant Professor with the Mechanical, Industrial and Systems Engineering Department and the Director of the Intelligent
Control and Robotics Laboratory, University of Rhode Island, Kingston, RI, U.S.A..
His research interests include adaptive learning and control, hybrid systems, and multirobot-distributed control.
 \end{IEEEbiography}

\begin{IEEEbiography}[{\includegraphics[width=1.0in,height=1.2in,clip]{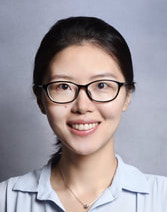}}]{Yan Gu}
received the B.S. degree in Mechanical Engineering from Zhejiang University, Hangzhou, China, in 2011 and the Ph.D. degree in Mechanical Engineering from Purdue University, West Lafayette, IN, U.S.A., in 2017.
She joined the faculty of the School of Mechanical Engineering at Purdue University in 2022.
Prior to joining Purdue, she was with the Department of Mechanical Engineering at the University of Massachusetts Lowell.
Her research interests include nonlinear control, hybrid systems, legged locomotion, and wearable robots.
She received the NSF CAREER Award in 2021.
\end{IEEEbiography}

\ifCLASSOPTIONcaptionsoff
  \newpage
\fi

\end{document}